\documentclass[letterpaper]{article} % DO NOT CHANGE THIS
\usepackage{aaai2026}  % DO NOT CHANGE THIS
\nocopyright
\usepackage{times}  % DO NOT CHANGE THIS
\usepackage{helvet}  % DO NOT CHANGE THIS
\usepackage{courier}  % DO NOT CHANGE THIS
\usepackage[hyphens]{url}  % DO NOT CHANGE THIS
\usepackage{graphicx} % DO NOT CHANGE THIS
\usepackage{natbib}  % DO NOT CHANGE THIS AND DO NOT ADD ANY OPTIONS TO IT
\usepackage{caption} % DO NOT CHANGE THIS AND DO NOT ADD ANY OPTIONS TO IT
\usepackage{algorithm}
\usepackage{newfloat}
\usepackage{listings}
\DeclareCaptionStyle{ruled}{labelfont=normalfont,labelsep=colon,strut=off} % DO NOT CHANGE THIS
\floatstyle{ruled}
\newfloat{listing}{tb}{lst}{}
\floatname{listing}{Listing}
\usepackage{amsmath}
\usepackage{tikz}
\usepackage{subcaption}
\usepackage{algpseudocode}
\usepackage{mathtools}
\usepackage{booktabs}
\usepackage{multirow}
\usepackage{siunitx}

\title{Exploration Through Introspection: \\ A Self-Aware Reward Model}
\author{
    Michael Petrowski,
    Milica Ga{\v{s}}i{\'c},
}
\affiliations{
    Heinrich Heine University Düsseldorf, Germany\\
    m.petrowski@uni-duesseldorf.de, 
    gasic@hhu.de
}

\begin{document}

\maketitle

\begin{abstract}
Understanding how artificial agents model internal mental states is central to advancing Theory of Mind in AI. Evidence points to a unified system for self- and other-awareness. 
We explore this self-awareness by having reinforcement learning agents infer their own internal states in gridworld environments. 
Specifically, we introduce an introspective exploration component that is inspired by biological pain as a learning signal by utilizing a hidden Markov model to infer \mbox{``pain-belief''} from online observations. This signal is integrated into a subjective reward function to study how self-awareness affects the agent's learning abilities. Further, we use this computational framework to investigate the difference in performance between normal and chronic pain perception models.
Results show that introspective agents in general significantly outperform standard baseline agents and can replicate complex human-like behaviors.
\end{abstract}

% Uncomment the following to link to your code, datasets, an extended version or similar.
% You must keep this block between (not within) the abstract and the main body of the paper.
\begin{links}
    \link{Code}{https://github.com/m-petrowski/pain_rl}
    \link{Datasets}{https://doi.org/10.5281/zenodo.18036125}
    %\link{Extended version}{https://aaai.org/example/extended-version}
\end{links}

\section{Introduction}

A defining feature of human intelligence is the ability to attribute hidden mental states like beliefs to oneself and others, known as \emph{Theory of Mind} (ToM) \cite{Premack1978}. 
While recent ToM models in AI often emphasize inference of beliefs of others \cite{ToM4AI2025}, a closely related aspect to understanding others is a sophisticated understanding of oneself. 
\citeauthor{Happe2003}~\shortcite{Happe2003} argues that humans possess a unified system for thinking about mental states, whether directed at oneself or others. 
Inspired by this system in humans, equipping an artificial agent with the ability to model its own affective states might provide a form of self-awareness that is closely related to the understanding of similar states in others.
In biological agents, signals such as pain are a core component of adaptive learning \cite{Seymour2019, Tabor2019}.
They encode internal evaluations that guide the agent through complex environments.

In this work, we propose a computational framework for introspection that provides an exploration incentive through self-evaluation to a reinforcement learning (RL) agent. Inspired by the exploratory properties of biological pain, we operationalize an aversive signal not as direct environmental feedback, but as a latent state an agent must infer.
Specifically, the agent's own internal belief state of pain is inferred from observations of its own ``happiness'', which is a reward signal proposed by \citeauthor{Dubey2022}~\shortcite{Dubey2022}.

Our novel reward acts as a powerful intrinsic learning signal that provides a dynamic exploration incentive and adapts to environmental changes.
Further, this introspective model enables us to study the performance of agents with different types of perceptions. We model normal and chronic pain perception and show how such a maladaptive perception can produce complex, addiction-like behaviors within our framework.

\section{Related Work}
Traditional RL frameworks optimize externally defined reward functions \cite{SuttonBarto2018}, lacking the representational depth needed for mental-state reasoning. 
Recent work introduces subjective reward functions inspired by human features, where agents evaluate outcomes relative to prior expectations and comparisons \cite{Dubey2022}. 
Similarly, Bayesian models of pain conceptualize pain perception as inference over hidden nociceptive causes \cite{Eckert2022}.
In cognitive science, ToM is often modeled through partially observable inference, where agents estimate others’ beliefs under uncertainty \cite{Baker2017}. 
The same probabilistic architecture applies to introspective inference: \citeauthor{Mahajan2025}~\shortcite{Mahajan2025} cast pain and injury as Partially Observable Markov Decision Processes. 
Our work builds on this paradigm by embedding an inferential affective model inside an RL agent, thereby merging Bayesian inference with decision-theoretic learning.

\section{A Self-Aware Reward Model}
Our introspective exploration component consists of a hidden Markov model (HMM) inspired by \citeauthor{Eckert2022}~\shortcite{Eckert2022}. We utilize the parameters from their HMMs that resemble normal and chronic pain perception (see Appendix~\ref{app:hmm_parameters}). Both models use hidden states 
${H_t\in H = \{\textup{pain}, \textup{no\_pain}\}}$ 
and observations ${O_t \in O = \{\textup{noxious}, \textup{harmless}\}}$.
Their chronic pain model has sticky transitions and ambiguous emissions, representing a maladaptive perception model. The normal pain model is more informative and represents the healthy counterpart. Henceforth, we will refer to our models as ``normal pain'' and ``chronic pain'' for convenience, acknowledging that they do not represent actual clinical pain in our framework.

We extend the happiness function $f^h$, which balances objective reward, expectations, and relative comparisons \cite{Dubey2022}, with our HMM-based aversive signal. The resulting function $f^w$ represents the agent's ``well-being'' and acts as the subjective reward function:
\begin{equation} \label{eq:happiness_function}
    \begin{aligned}
    f^h &= w_1 \cdot \textup{Objective} + w_2 \cdot \textup{Expect} + w_3 \cdot \textup{Compare} \\
    %%%%&= w_1 \cdot r_{t+1} + w_2(r_{t+1} - Q(s_t,a_t)) + w_3(r_{t+1} - \rho) \\
    f^w &= f^h - w_4 \cdot \textup{Pain}
    \end{aligned} 
\end{equation}
For learning, we use the Q-Learning algorithm \cite{Watkins1992} with $f^w$ as the reward function in combination with an $\epsilon$-greedy policy.
The $\textup{Pain}$ component from $f^w$ is defined as the belief state of pain:
\begin{equation} \label{eq:belief_state}
        \textup{Pain} \doteq b(H_t =\textup{pain}) = \Pr(H_t = \textup{pain}\mid O_{1:t})
\end{equation}
The beliefs are updated online using the forward algorithm \cite{Rabiner1989}. 
The observations are obtained by incorporating the sensory information given by $f^h$. 
At every time step $t$, after taking action $a_t$ in state $s_t$ and transitioning into state $s_{t+1}$, we observe the value of the happiness $f^h$. If $f^h \geq 0$, then $O_t$ is $\textup{harmless}$ and if $f^h < 0$, it is $\textup{noxious}$.

\section{Experiments \& Results}\label{sec:experiments_results}

We evaluate agents with introspective exploration in experiments across two different gridworld environments. 
Each is a $7 \times 7$ gridworld that contains a single agent and a food state (+1 objective reward, all other states 0), where the goal is to reach the food as quickly and as often as possible.
In the \emph{stationary} environment, the food state does not change its position throughout the agent's lifetime (2500 steps).
In the \emph{non-stationary} environment, the food state changes its location to one of the other corners after every 1250 time steps during the agent's lifetime (5000 steps).
The setup is inspired by \citeauthor{Dubey2022}~\shortcite{Dubey2022}.

For each environment, we perform a grid search to find the parameters of the best performing agents with normal and chronic pain models from various reward function groups using the optimal reward framework \cite{Singh2009}, with the mean cumulative objective reward (COR) as the performance metric. 
Reward function groups are determined by the components of $f^h$ that are activated via their weights $w > 0$ (see Appendix~\ref{app:experiment_details} for details).

\begin{figure}[t!]
    \centering
    \includegraphics[width=1\columnwidth]{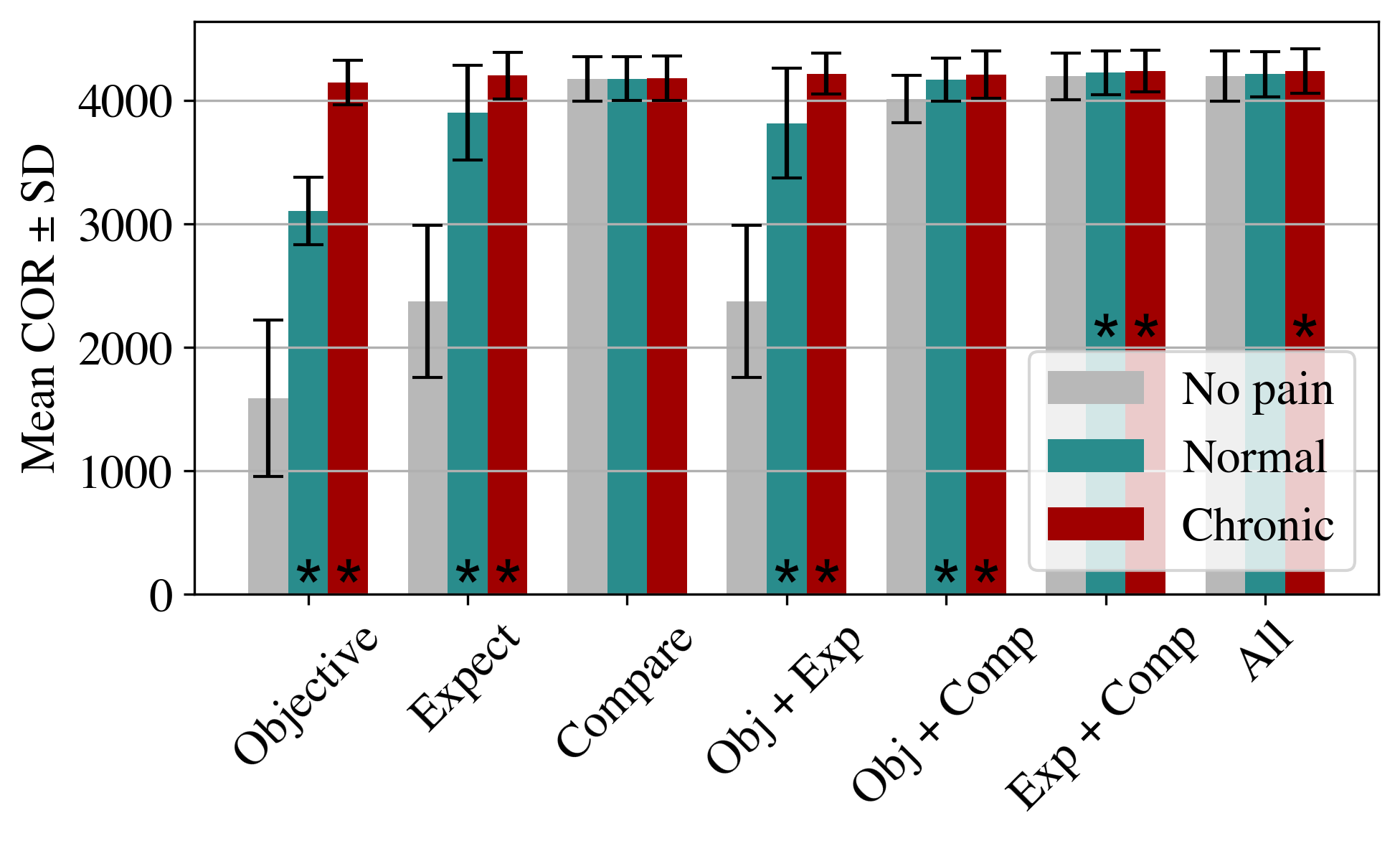}
    \caption{Non-stationary: Mean cumulative objective reward (COR) and standard deviation (SD) of the best performing agents from each reward function category across different self-awareness models. (*): Statistically significant improvement over the same category `No pain' baseline (one-sided paired-samples t-test, $p \ll 0.05$). Sample size ${n=300}$.}
    \label{fig:results_non_stationary}
\end{figure}

Our results show that introspective agents generally outperform `No pain' baseline agents in their respective reward category across environments (Figure~\ref{fig:results_non_stationary}, Figure~\ref{fig:results_stationary}). This illustrates faster adaptation while receiving a dynamic exploration incentive through the aversive signal (Figure~\ref{fig:results_non_stationary}, Appendix~\ref{app:detailed_results}).
We also observed that agents with normal (${M=2295.6}, {SD=65.7}$) and chronic (${M=2295.0}, {SD=66.1}$) pain performed similarly in the stationary `Objective+Expect' category, whereas the chronic model (${M=4214.6}, {SD=165.4}$) interestingly outperformed the normal one (${M=3814.0}, {SD=446.6}$) in the non-stationary setting. In both scenarios, the agents using self-awareness were superior to their respective `No pain' baselines% (Stationary: ${M=1973.1}, {SD=385.0}$; Non-stationary: ${M=2371.0}, {SD=613.3}$)
.
This also applies to other categories (Figure~\ref{fig:results_non_stationary}, Appendix~\ref{app:detailed_results}).

In the non-stationary setting, normal (${\alpha = 0.9}$, `Objective only') and chronic (${\alpha = 0.1}$, `Objective only') agents perform better with opposite learning rates $\alpha$, suggesting differences in learning between perception models. Interestingly, the chronic pain agent notably outperforms its counterpart at the cost of a negative cumulative well-being throughout its lifetime. Its momentary well-being reaches zero only when visiting the food state, paralleling relief-seeking behavior (Appendix~\ref{app:further_discussion}). However, agents with this maladaptive component also yield the worst overall performance within the `Objective+Expect' category (Appendix~\ref{app:detailed_results}, Figure~\ref{fig:reward_distribution_non_stationary}).

\section{Discussion \& Conclusion}
Our introspection component serves as a powerful tool for adaptive learning: the normal model acts as a low-pass filter, smoothing the happiness signal $f^h$ into a stable belief state that provides a dynamic exploration bonus, which is particularly advantageous in non-stationary environments.
The chronic model, while approximating a constant decrement similar to the fixed aspiration level in \citeauthor{Dubey2022}~\shortcite{Dubey2022}, is valuable for the complex behaviors it produces. When given a psychological interpretation, the agent's high performance despite negative well-being aligns with findings on chronic pain and quality of life \cite{Gureje1998, Blyth2001}. Furthermore, its relief-seeking behavior through faster recovery provides a computational parallel to negative reinforcement in addiction \cite{Koob2008}. For more details, see Appendix~\ref{app:further_discussion}.

While this work has limitations, such as using a simple $\epsilon$-greedy baseline and fixed HMM parameters, it successfully demonstrates that an RL agent's performance improves when it models one of its own internal aversive states. This self-awareness enhances adaptation and can generate complex, psychologically plausible dynamics. Our framework models the self-application aspect of the unified system hypothesized for ToM, and future work can test this by extending the architecture to infer others' states.

\bibliography{aaai2026}

\appendix

\section{Hidden Markov Model Parameters}\label{app:hmm_parameters}
Parameters for the HMMs can be found in Figure~\ref{fig:normal_hmm} and Figure~\ref{fig:chronic_hmm}.

\begin{figure}[h!]
    \centering
    \resizebox{0.6\columnwidth}{!}{
    \begin{tikzpicture}
\tikzstyle{h_state}=[circle, draw=black, minimum size=0.75cm, inner sep=0pt]
\tikzstyle{o_state}=[circle, draw=black, fill=black!20, minimum size=0.75cm, inner sep=0pt]
\tikzstyle{transition}=[->, thick, black]

  \node[draw, rectangle, inner sep=5pt] (P) at (0,-1.7){
    \begin{tabular}{c|c|c}
      $H_{t-1}|H_t$ & $\textup{pain}$ & $\textup{no\_pain}$ \\
      \hline
      $\textup{pain}_{t-1}$ & 0.3 & 0.7 \\
      $\textup{no\_pain}_{t-1}$ & 0.2 & 0.8 \\
    \end{tabular}
  };

\node[draw, rectangle, inner sep=5pt] (E) at (0,-3.4){
    \begin{tabular}{c|c|c}
      $H_t|O_t$ & $\textup{noxious}$ & $\textup{harmless}$ \\
      \hline
      $\textup{pain}$ & 0.8 & 0.2 \\
      $\textup{no\_pain}$ & 0.1 & 0.9 \\
    \end{tabular}
  };

  \node[draw, rectangle, inner sep=5pt] (pi) at (0,0){
    \begin{tabular}{c|c}
      $H_0$ &\\
      \hline
      $\textup{pain}$ & 0.223 \\
      $\textup{no\_pain}$ & 0.777 \\
    \end{tabular}
  };

\end{tikzpicture}
}
    \caption{
        Normal pain: Parameters of the transition matrix ${\Pr(H_t \mid H_{t-1})}$, emission matrix ${\Pr(O_t \mid H_t)}$, and initial state distribution ${\Pr(H_0)}$ of the hidden Markov model for normal pain perception. Transitions favor recovery, and emissions distinguish noxious from harmless sensations. Adapted from \citeauthor{Eckert2022}~\shortcite{Eckert2022}.
}
\label{fig:normal_hmm}
\end{figure}

\begin{figure}[h!]
    \centering
    \resizebox{0.6\columnwidth}{!}{
        \begin{tikzpicture}
\tikzstyle{h_state}=[circle, draw=black, minimum size=0.75cm, inner sep=0pt]
\tikzstyle{o_state}=[circle, draw=black, fill=black!20, minimum size=0.75cm, inner sep=0pt]
\tikzstyle{transition}=[->, thick, black]

  \node[draw, rectangle, inner sep=5pt] (P) at (0,-1.7){
    \begin{tabular}{c|c|c}
      $H_{t-1}|H_t$ & $\textup{pain}$ & $\textup{no\_pain}$ \\
      \hline
      $\textup{pain}_{t-1}$ & 0.8 & 0.2 \\
      $\textup{no\_pain}_{t-1}$ & 0.7 & 0.3 \\
    \end{tabular}
  };

\node[draw, rectangle, inner sep=5pt] (E) at (0,-3.4){
    \begin{tabular}{c|c|c}
      $H_t|O_t$ & $\textup{noxious}$ & $\textup{harmless}$ \\
      \hline
      $\textup{pain}$ & 0.6 & 0.4 \\
      $\textup{no\_pain}$ & 0.6 & 0.4 \\
    \end{tabular}
  };

  \node[draw, rectangle, inner sep=5pt] (pi) at (0,0){
    \begin{tabular}{c|c}
      $H_0$ &\\
      \hline
      $\textup{pain}$ & 0.777 \\
      $\textup{no\_pain}$ & 0.223 \\
    \end{tabular}
  };

\end{tikzpicture}
}
    \caption{Chronic pain: Parameters of the transition matrix ${\Pr(H_t \mid H_{t-1})}$, emission matrix ${\Pr(O_t \mid H_t)}$, and initial state distribution ${\Pr(H_0)}$ of the hidden Markov model for chronic pain perception. Transitions are sticky and emissions are ambiguous. Adapted from \citeauthor{Eckert2022}~\shortcite{Eckert2022}.}
    \label{fig:chronic_hmm}
\end{figure}

\section{Experiment Setup}\label{app:experiment_details}
\begin{figure}[h!]
    \centering
    \resizebox{0.6\columnwidth}{!}{
    \includegraphics[width=\linewidth]{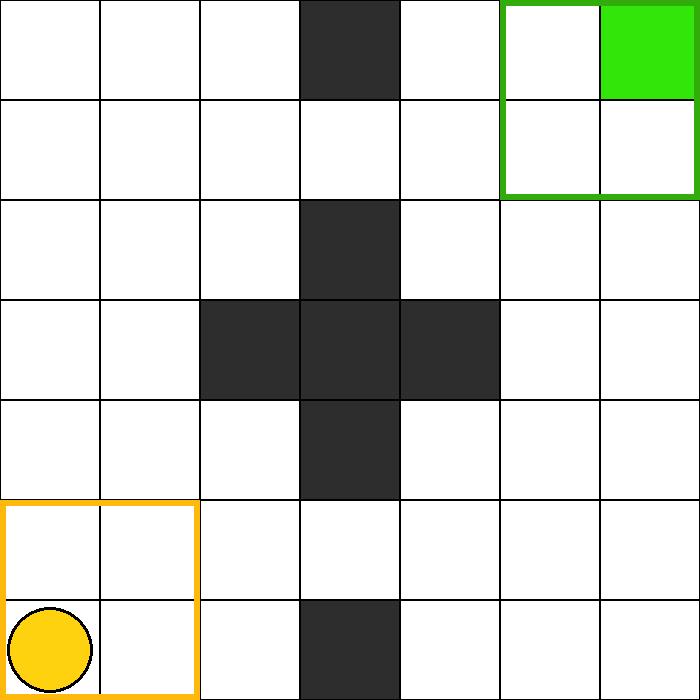}
    }
    \caption{Basic environment setup. Yellow circle: agent. Green square: food state. Yellow-marked area: possible initial agent spawn positions. Green-marked area: possible initial food state spawn positions.}
    \label{fig:basic_env_setup}
\end{figure}

All agents in the experiments use the Q-learning algorithm with an $\epsilon$-greedy policy to balance exploration and exploitation. The agent spawns in the bottom left corner and can navigate the environment by choosing one of five actions at each time step: \texttt{UP}, \texttt{DOWN}, \texttt{LEFT}, and \texttt{RIGHT} move the agent in the corresponding direction to the following field in the gridworld. The fifth action \texttt{STAY} makes it possible for the agent to keep its position on the current field. Figure~\ref{fig:basic_env_setup} shows this basic environment setup.

\subsection{Grid Search}
According to the optimal reward framework \cite{Singh2009}, we produce histories for all reward functions $r_A \in R_A$ and evaluate the histories using the fitness function $F$, which is the cumulative objective reward at the end of agent $A$'s lifetime. In our setup, the cumulative objective reward corresponds to the number of time steps the agent has visited the food state during its lifetime, as only the food state grants the agent an objective reward of $+1$.
Specifically, for each reward function $r_A$, we will produce $n=300$ histories, sampling a new environment $E \in \mathcal{E}_{\textup{stationary}}$ for each trajectory, and take the mean of $F(h)$ over all those $300$ histories. Also, the set of possible subjective reward functions is divided into what we call reward categories that correspond to the weights $w_1,w_2,w_3$ that are activated. 
There are seven of those categories: `Objective only', `Expect only', `Compare only', `Objective+Expect', `Objective+Compare', `Expect+Compare' and `All'. Each category is then divided into three subcategories, corresponding to the used pain model, which are \texttt{No pain}, \texttt{Normal pain} and \texttt{Chronic pain}. 
For each reward subcategory, we determine the best-performing agent according to the optimal reward function $r^*_A$ using the grid search. Because the grid search includes a large number of hyperparameters and grows exponentially with them, we limited our search space to $w_1,w_2,w_3,w_4 \in [0, 0.1, 0.3, 0.5, 0.7, 0.9, 1]$, the aspiration level $\rho \in [0.01, 0.05, 0.1, 0.3, 0.5, 0.7, 0.9, 1]$, the learning rate $\alpha \in [0.1, 0.3, 0.5, 0.7, 0.9]$ and the exploration rate $\epsilon \in [0.01, 0.1]$ to save computational resources, while also including the two HMMs for normal and chronic pain. Removing duplicate or invalid reward functions results in a total of $312{,}130$ subjective reward functions for each of the two environments. The used discount factor for all experiments was $\gamma = 0.99$.

\section{Detailed Results}\label{app:detailed_results}

\subsection{Stationary Environment}

\begin{figure}[h!]
    \centering
    \includegraphics[width=1\columnwidth]{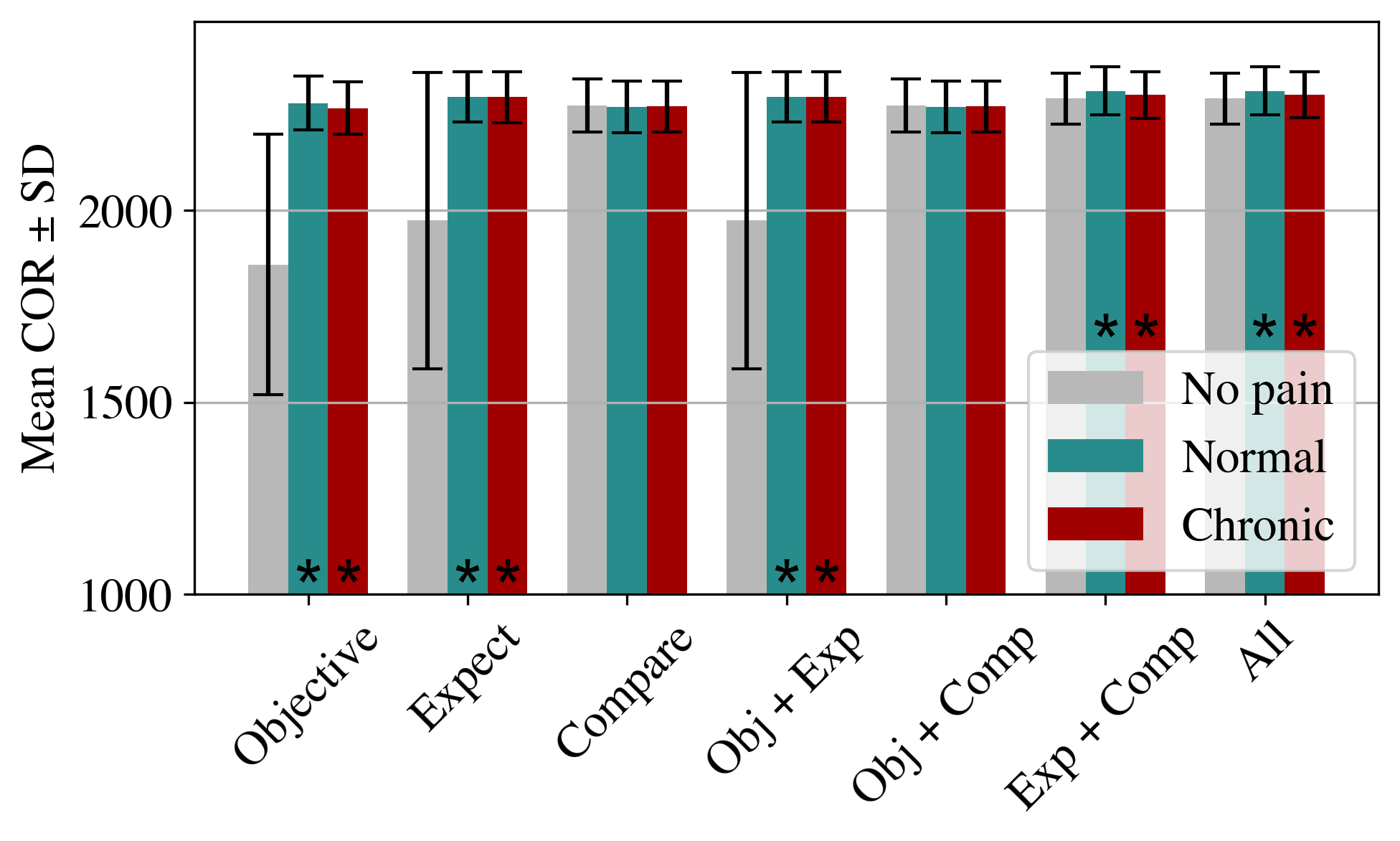}
    \caption{Stationary environment: Mean cumulative objective reward (COR) and standard deviation (SD) of the best performing agents from each reward function category across different self-awareness models. (*): Statistically significant improvement over the same category `No pain' baseline (one-sided paired-samples t-test, $p \ll 0.05$). Sample size $n=300$.
    To highlight the differences, the plot starts from 1000 on the y-axis.}
    \label{fig:results_stationary}
\end{figure}

Detailed results of the experiments in the stationary environment can be taken from Figure~\ref{fig:results_stationary} and Table~\ref{tab:results_stationary}.

\begin{table}[ht!]
    \centering
    {
    \small
    \setlength{\tabcolsep}{1mm}
    \begin{tabular}{cccccccccS[table-format=4.1]c}
         \toprule
         \multicolumn{11}{c}{\textbf{Stationary Environment Results}} \\
         \midrule
         \textbf{Cat.} & \textbf{Pain} & \textbf{$w_1$} & \textbf{$w_2$} & \textbf{$w_3$} & \textbf{$w_4$} & \textbf{$\rho$} & \textbf{$\epsilon$} & \textbf{$\alpha$} & \textbf{Mean} & \textbf{SD}\\
         \midrule
         \multirow{3}{*}{Obj.} 
             &/& $0.9$ & $0.0$ & $0.0$ & $0.0$ & \textsc{{NA}} & $.1$ & $.9$ & 1858.6 & $338.7$ \\
             &N& $0.1$ & $0.0$ & $0.0$ & $1.0$ & \textsc{{NA}} & $.01$ & $.7$ & 2279.5$^*$ & $69.3$ \\
             &C& $0.1$ & $0.0$ & $0.0$ & $0.1$ & \textsc{{NA}} & $.01$ & $.9$ & 2266.3$^*$ & $68.5$ \\
         \midrule
         \multirow{3}{*}{Exp.} 
             &/& $0.0$ & $0.9$ & $0.0$ & $0.0$ & \textsc{{NA}} & $.01$ & $.7$ & 1973.1 & $385.0$ \\
             &N& $0.0$ & $0.7$ & $0.0$ & $0.7$ & \textsc{{NA}} & $.01$ & $.7$ & 2295.6$^*$ & $65.7$ \\
             &C& $0.0$ & $0.3$ & $0.0$ & $0.3$ & \textsc{{NA}} & $.01$ & $.9$ & 2294.6$^*$ & $66.0$ \\
         \midrule
         \multirow{3}{*}{Comp.} 
             &/& $0.0$ & $0.0$ & $0.9$ & $0.0$ & $0.7$ & $.01$ & $.9$ & 2272.2 & $69.1$ \\
             &N& $0.0$ & $0.0$ & $0.5$ & $0.9$ & $0.05$ & $.01$ & $.7$ & 2269.3 & $67.7$ \\
             &C& $0.0$ & $0.0$ & $1.0$ & $0.1$ & $0.7$ & $.01$ & $.9$ & 2270.2 & $67.0$ \\
         \midrule
         \multirow{3}{*}{\shortstack{Obj. +\\Exp.}} 
             &/& $0.5$ & $0.9$ & $0.0$ & $0.0$ & \textsc{{NA}} & $.01$ & $.7$ & 1973.1 & $385.0$ \\
             &N& $0.1$ & $0.7$ & $0.0$ & $0.7$ & \textsc{{NA}} & $.01$ & $.7$ & 2295.6$^*$ & $65.7$ \\
             &C& $0.9$ & $0.3$ & $0.0$ & $0.7$ & \textsc{{NA}} & $.01$ & $.9$ & 2295.0$^*$ & $66.1$ \\
         \midrule
         \multirow{3}{*}{\shortstack{Obj. +\\Comp.}} 
             &/& $0.1$ & $0.0$ & $0.5$ & $0.0$ & $1.0$ & $.01$ & $.7$ & 2272.2 & $69.1$ \\
             &N& $0.1$ & $0.0$ & $0.5$ & $0.9$ & $0.05$ & $.01$ & $.7$ & 2269.3 & $67.7$ \\
             &C& $0.1$ & $0.0$ & $0.5$ & $0.1$ & $1.0$ & $.01$ & $.7$ & 2270.3 & $67.0$ \\
         \midrule
         \multirow{3}{*}{\shortstack{Exp. +\\Comp.}} 
             &/& $0.0$ & $0.7$ & $1.0$ & $0.0$ & $0.9$ & $.01$ & $.7$ & 2291.1 & $65.8$ \\
             &N& $0.0$ & $0.3$ & $0.3$ & $0.7$ & $0.01$ & $.01$ & $.1$ & 2310.6$^*$ & $62.5$ \\
             &C& $0.0$ & $0.7$ & $0.3$ & $0.5$ & $0.05$ & $.01$ & $.7$ & 2300.4$^*$ & $61.0$ \\
         \midrule
         \multirow{3}{*}{All} 
             &/& $0.1$ & $0.7$ & $0.7$ & $0.0$ & $1.0$ & $.01$ & $.7$ & 2291.0 & $65.8$ \\
             &N& $0.7$ & $0.3$ & $0.3$ & $0.7$ & $0.01$ & $.01$ & $.1$ & 2310.6$^*$ & $62.5$ \\
             &C& $0.7$ & $0.3$ & $0.7$ & $1.0$ & $0.01$ & $.01$ & $.9$ & 2300.6$^*$ & $59.6$ \\
         \bottomrule
    \end{tabular}
    }
    \caption{Stationary Environment: Parameters and corresponding mean cumulative objective reward (Mean) and standard deviation (SD) of the best performing agents from each reward function category across different pain conditions in the stationary environment. The reward categories are divided into pain subcategories: `No pain' (/), `Normal pain' (N) and `Chronic pain' (C). Results marked with an asterisk (*) indicate a statistically significant improvement over the `No pain' baseline within the same category (one-sided paired-samples t-test, $p \ll 0.05$). Mean and SD are rounded to one decimal place. $\text{Lifetime} = 2500$, sample size $n=300$.}
    \label{tab:results_stationary}
\end{table}

\subsection{Non-Stationary Environment}

\begin{figure}[h!]
    \centering
    \includegraphics[width=\columnwidth]{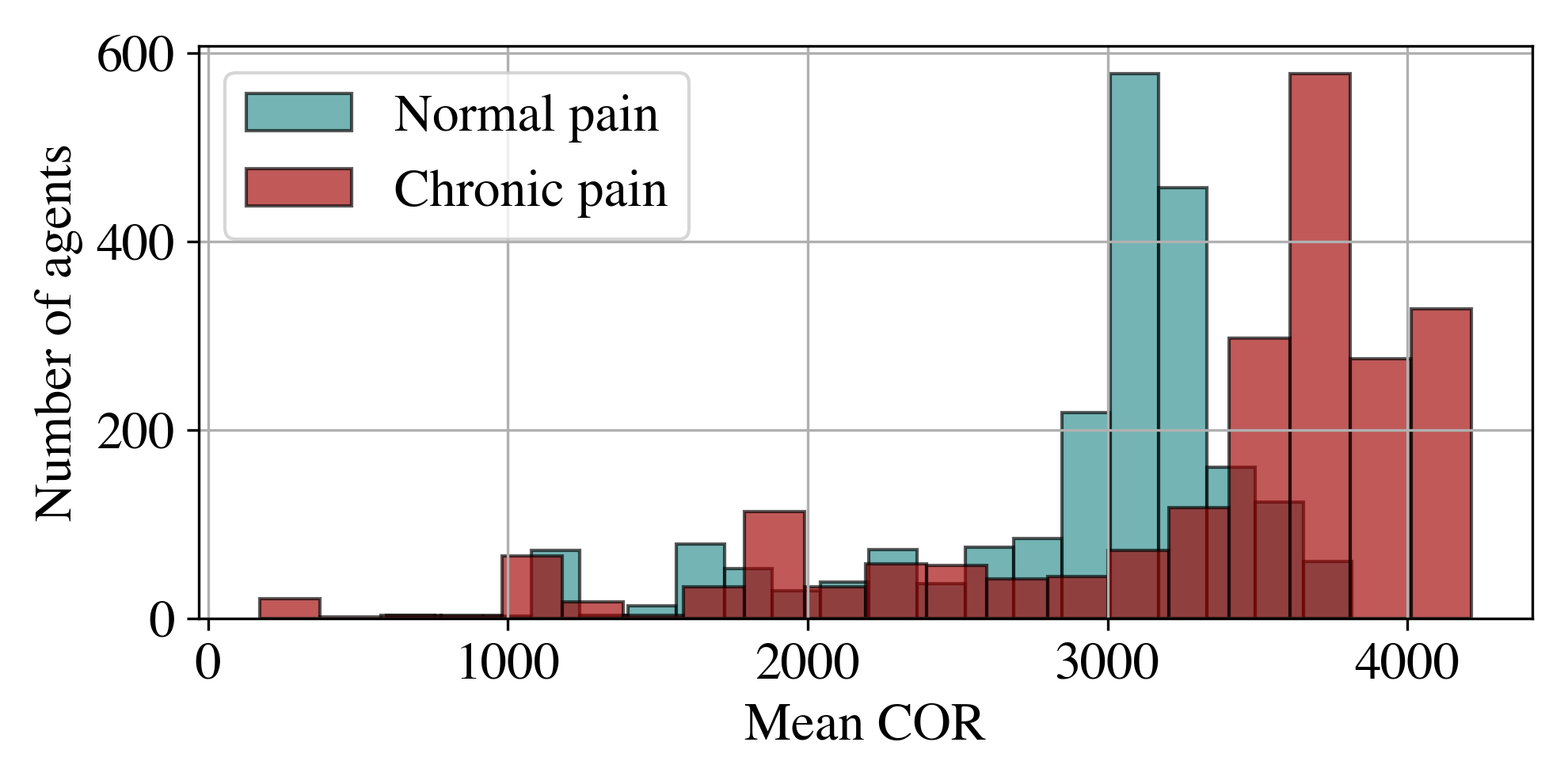}
    \caption{Non-stationary environment: Distribution of mean cumulative objective reward (COR) in the `Objective+Expect' category, comparing all reward functions of the agents from the grid search with normal vs. chronic pain perception in this category.}
    \label{fig:reward_distribution_non_stationary}
\end{figure}

\begin{figure}[h!]
    \centering
    \includegraphics[width=1\columnwidth]{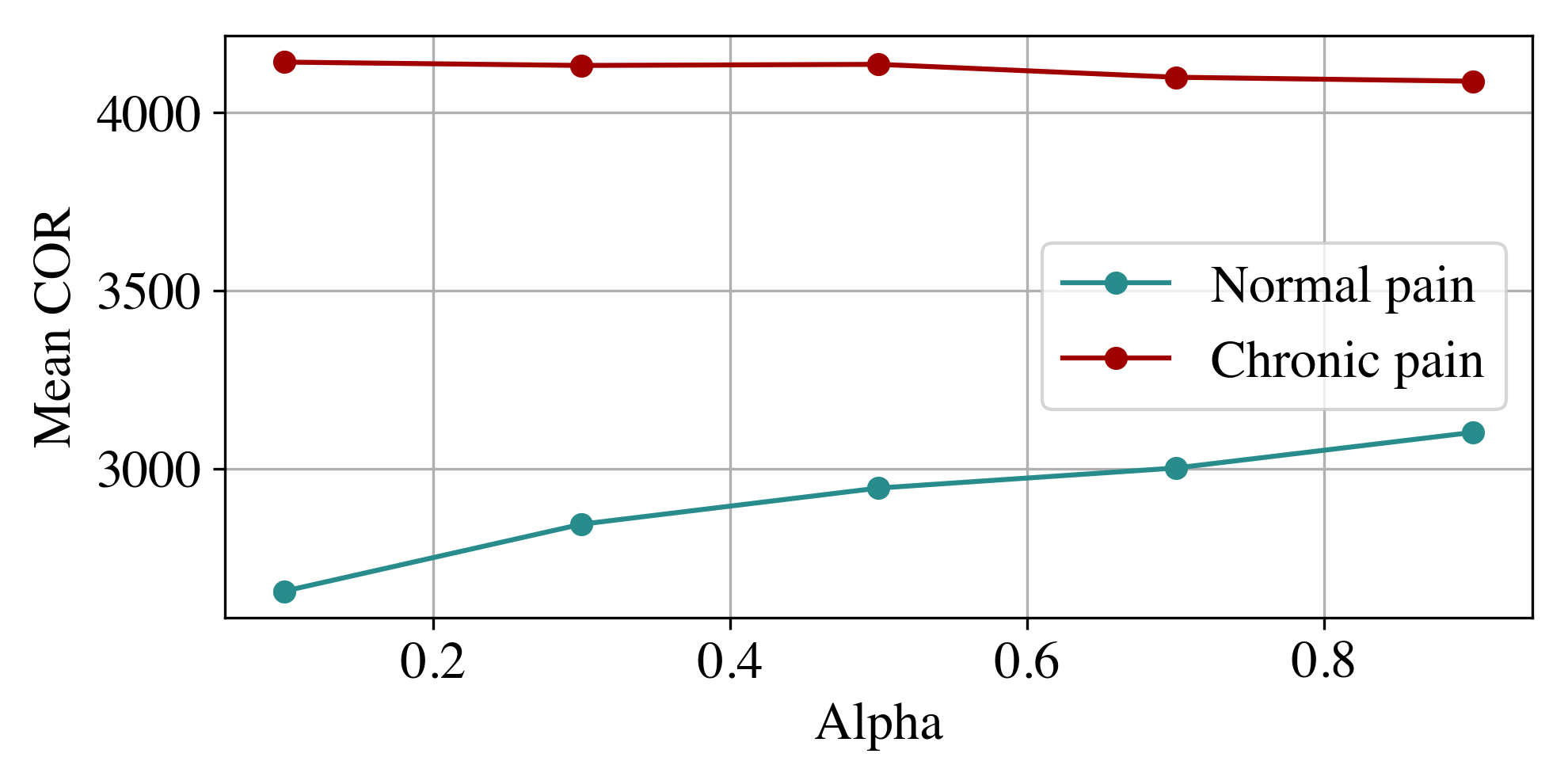}
    \caption{Non-stationary environment: Mean cumulative objective reward (COR) of respective best-performing `Objective only' agents with normal vs. chronic pain perception, plotted across different learning rates $\alpha$.}
    \label{fig:alphas_non_stationary}
\end{figure}

\begin{figure}[h!]
    \centering
    \includegraphics[width=\linewidth]{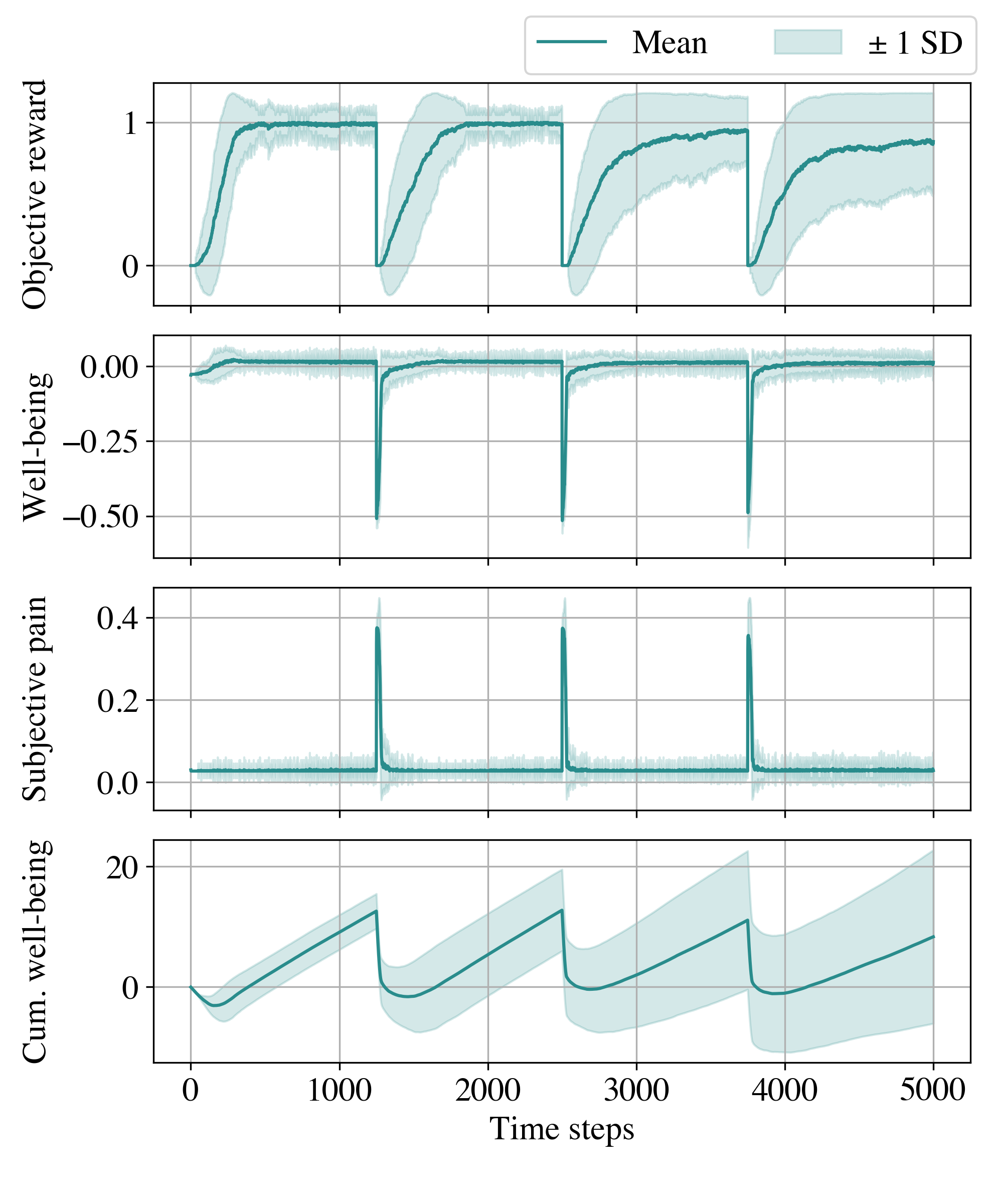}
    \caption{Normal pain, non-stationary environment: Mean and standard deviation (SD) of the Objective reward, momentary well-being, subjective pain ($w_4 \cdot \textup{Pain}$) and cumulative well-being of the best performing `Objective+Expect, Normal pain' agent, plotted over the agent's lifetime (5000 steps).}
    \label{fig:np_metrics_non_stationary}
\end{figure}

\begin{figure}[h!]
    \centering
    \includegraphics[width=\linewidth]{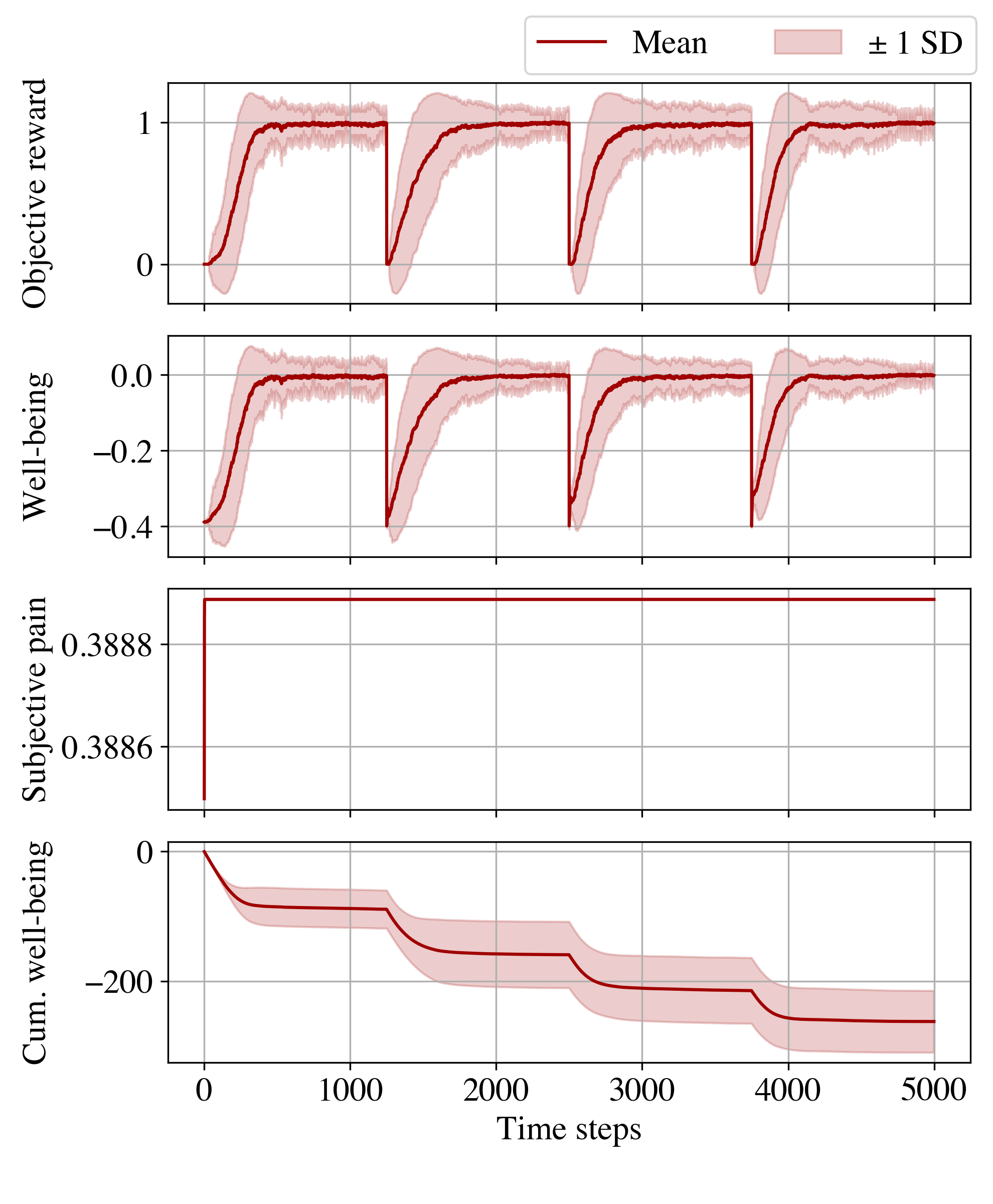}
    \caption{Chronic pain, non-stationary environment: Mean and standard deviation (SD) of the Objective reward, momentary well-being, subjective pain ($w_4 \cdot \textup{Pain}$) and cumulative well-being of the best performing `Objective+Expect, Chronic pain' agent, plotted over the agent's lifetime (5000 steps).}
    \label{fig:cp_metrics_non_stationary}
\end{figure}

Detailed results of the experiments in the non-stationary environment can be taken from Figure~\ref{fig:results_non_stationary} and Table~\ref{tab:results_non_stationary}.
Figure~\ref{fig:alphas_non_stationary} shows the performance of normal and chronic agents in the `Objective only' category across different learning {rates $\alpha$} (see Section~\ref{sec:experiments_results}).
Plots of the objective reward, momentary well-being, subjective pain ($w_4 \cdot \textup{Pain}$) and cumulative well-being of the best normal and chronic pain agents in the `Objective+Expect' category can be found in Figure~\ref{fig:np_metrics_non_stationary} and Figure~\ref{fig:cp_metrics_non_stationary}, respectively.
Figure~\ref{fig:reward_distribution_non_stationary} shows the distribution of the performance of all the normal and chronic agents in the `Objective+Expect' category that were part of the grid search, mentioned in Section~\ref{sec:experiments_results}.

\begin{table}[ht!]
    \centering
    {
    \small
    \setlength{\tabcolsep}{1mm}
    \begin{tabular}{cccccccccS[table-format=4.1]c}
         \toprule
         \multicolumn{11}{c}{\textbf{Non-Stationary Environment Results}} \\
         \midrule
         \textbf{Cat.} & \textbf{Pain} & \textbf{$w_1$} & \textbf{$w_2$} & \textbf{$w_3$} & \textbf{$w_4$} & \textbf{$\rho$} & \textbf{$\epsilon$} & \textbf{$\alpha$} & \textbf{Mean} & \textbf{SD}\\
         \midrule
         \multirow{3}{*}{Obj.} 
         &/& $0.1$ & $0.0$ & $0.0$ & $0.0$ & \textsc{{NA}} & $.1$ & $.9$ & 1586.5 & $631.2$ \\
         &N& $0.1$ & $0.0$ & $0.0$ & $1.0$ & \textsc{{NA}} & $.01$ & $.9$ & 3101.8$^*$ & $271.8$ \\
         &C& $0.7$ & $0.0$ & $0.0$ & $0.9$ & \textsc{{NA}} & $.01$ & $.1$ & 4142.5$^*$ & $177.2$ \\
         
         \midrule
         \multirow{3}{*}{Exp.} 
         &/& $0.0$ & $1.0$ & $0.0$ & $0.0$ & \textsc{{NA}} & $.1$ & $.7$ & 2371.0 & $613.3$ \\
         &N& $0.0$ & $0.1$ & $0.0$ & $0.9$ & \textsc{{NA}} & $.01$ & $.1$ & 3896.3$^*$ & $383.1$ \\
         &C& $0.0$ & $0.3$ & $0.0$ & $0.3$ & \textsc{{NA}} & $.01$ & $.3$ & 4197.7$^*$ & $186.9$ \\

         \midrule
         \multirow{3}{*}{Comp.} 
         &/& $0.0$ & $0.0$ & $0.1$ & $0.0$ & $1.0$ & $.01$ & $.1$ & 4171.1 & $178.9$ \\
         &N& $0.0$ & $0.0$ & $0.9$ & $1.0$ & $0.9$ & $.01$ & $.3$ & 4173.1 & $178.0$ \\
         &C& $0.0$ & $0.0$ & $0.9$ & $0.1$ & $0.9$ & $001$ & $.3$ & 4178.1 & $181.6$ \\

         \midrule
         \multirow{3}{*}{\shortstack{Obj. +\\Exp.}}
         &/& $0.5$ & $1.0$ & $0.0$ & $0.0$ & \textsc{{NA}} & $.1$ & $.7$ & 2371.0 & $613.3$ \\
         &N& $0.1$ & $0.1$ & $0.0$ & $0.5$ & \textsc{{NA}} & $.01$ & $.7$ & 3814.0$^*$ & $446.6$ \\
         &C& $0.1$ & $0.3$ & $0.0$ & $0.5$ & \textsc{{NA}} & $.01$ & $.3$ & 4214.6$^*$ & $165.4$ \\

         \midrule
         \multirow{3}{*}{\shortstack{Obj. +\\Comp.}} 
         &/& $0.1$ & $0.0$ & $1.0$ & $0.0$ & $1.0$ & $.01$ & $.3$ & 4008.8 & $189.1$ \\
         &N& $0.1$ & $0.0$ & $1.0$ & $1.0$ & $1.0$ & $.01$ & $.3$ & 4165.9$^*$ & $172.2$ \\
         &C& $0.3$ & $0.0$ & $0.3$ & $0.5$ & $0.7$ & $.01$ & $.3$ & 4205.3$^*$ & $190.8$ \\

         \midrule
         \multirow{3}{*}{\shortstack{Exp. +\\Comp.}} 
         &/& $0.0$ & $0.1$ & $1.0$ & $0.0$ & $1.0$ & $.01$ & $.3$ & 4194.2 & $188.8$ \\
         &N& $0.0$ & $0.1$ & $1.0$ & $0.9$ & $1.0$ & $.01$ & $.3$ & 4222.5$^*$ & $176.0$ \\
         &C& $0.0$ & $0.3$ & $0.3$ & $0.5$ & $0.7$ & $.01$ & $.1$ & 4235.3$^*$ & $170.0$ \\

         \midrule
         \multirow{3}{*}{All} 
         &/& $0.1$ & $0.5$ & $1.0$ & $0.0$ & $1.0$ & $.01$ & $.3$ & 4194.6 & $201.2$ \\
         &N& $0.1$ & $0.3$ & $1.0$ & $0.7$ & $1.0$ & $.01$ & $.3$ & 4210.0 & $184.4$ \\
         &C& $0.1$ & $0.7$ & $0.7$ & $1.0$ & $1.0$ & $.01$ & $.1$ & 4235.5$^*$ & $180.3$ \\

         \bottomrule
    \end{tabular}
    }
    \caption{Non-stationary environment: Parameters and corresponding mean cumulative objective reward (Mean) and standard deviation (SD) of the best performing agents from each reward function category across different pain conditions in the non-stationary environment. The reward categories are divided into pain subcategories: `No pain' (/), `Normal pain' (N) and `Chronic pain' (C). Results marked with an asterisk (*) indicate a statistically significant improvement over the `No pain' baseline within the same category (one-sided paired-samples t-test, $p \ll 0.05$). Mean and SD are rounded to one decimal place. $\text{Lifetime} = 5000$, sample size $n=300$.}
    \label{tab:results_non_stationary}
\end{table}

%\clearpage
\section{Further Analysis}\label{app:further_discussion}
In the non-stationary environment, the chronic agent adapts faster to changes than the normal agent. Yet, the former consistently accumulates a negative cumulative well-being across its lifetime, while the normal agent remains positive. For the chronic pain agent, momentary well-being only recovers to approximately zero from a negative state when the food state is reached, as can be seen in the nearly identical trajectories of both objective reward and well-being in Figure~\ref{fig:cp_metrics_non_stationary}. 
This means that food discovery provides only temporary relief rather than a sustained positive value \cite{Navratilova2012}. The pattern strongly resembles how painkillers or addictive substances work in humans: consuming the drug brings short-term relief from a negative baseline, but once the effect wears off, well-being quickly drops back into the negative domain \cite{Koob2008}. For the chronic pain agent, being on the food state results in a neutral (zero) well-being, but losing access to the food causes an immediate drop below zero due to the high, persistent pain signal. This negative state is only alleviated by finding the food again, creating a cycle of relief-seeking behavior. 
Paradoxically, this need to escape the negative baseline improves task performance compared to the normal agent. The normal pain agent, in contrast, does not exhibit this cycle. Its momentary well-being briefly dips below zero when the food location changes, but then recovers to a non-negative level even before the food is found, leading to a stable and mostly positive cumulative well-being over its lifetime. \\

\end{document}